\documentclass[conference]{IEEEtran}
\ifCLASSINFOpdf
\else
\fi

\usepackage{graphicx}
\usepackage{bbold}
\usepackage{amssymb}
\usepackage{amsmath}
\usepackage[]{algorithmic}
\usepackage{algorithm}
\usepackage{multirow}
\usepackage{hhline}
\usepackage{balance}
\usepackage[font=small]{caption}
\usepackage{adjustbox}
\usepackage{balance}
\usepackage{amsthm}
\usepackage{cite}
\newtheorem{theorem}{Theorem}[section]

\begin{document}
%
\title{Multi-Source Multi-View Clustering via Discrepancy Penalty}



%
\author{
\IEEEauthorblockN{Weixiang Shao\IEEEauthorrefmark{1},
Jiawei Zhang\IEEEauthorrefmark{1},
Lifang He\IEEEauthorrefmark{2},
Philip S. Yu\IEEEauthorrefmark{1}
}
\IEEEauthorblockA{\IEEEauthorrefmark{1}University of Illinois at Chicago\\
Email: wshao4@uic.edu,
jzhan9@uic.edu,
psyu@uic.edu}
\IEEEauthorblockA{\IEEEauthorrefmark{2}Shenzhen University, China\\
Email: lifanghescut@gmail.com}
}


\maketitle
\begin{abstract}
With the advance of technology,
entities can be observed in multiple views.
Multiple views containing different types of features can be used for clustering.
Although multi-view clustering has been successfully applied in many applications,
the previous methods usually assume the complete instance mapping between different views.
In many real-world applications,
information can be gathered from multiple sources,
while each source can contain multiple views, which are more cohesive for learning.
The views under the same source are usually fully mapped, but they can be very heterogeneous.
Moreover, the mappings between different sources are usually incomplete and partially observed,
which makes it more difficult to integrate all the views across different sources.
In this paper, we propose MMC (\textbf{M}ulti-source \textbf{M}ulti-view \textbf{C}lustering),
which is a framework based on collective spectral clustering with a discrepancy penalty across sources,
to tackle these challenges.
MMC has several advantages compared with other existing methods.
First, MMC can deal with incomplete mapping between sources.
Second, it considers the disagreements between sources
while treating views in the same source as a cohesive set.
Third, MMC also tries to infer the instance similarities across sources to enhance the clustering performance.
Extensive  experiments  conducted  on real-world data demonstrate the effectiveness of the proposed approach.
\end{abstract}


%
\IEEEpeerreviewmaketitle

\section{Introduction}
With the advance of technology,
most of entities can be observed in multiple views.
Multiple views containing different types of features can be used for clustering.
Multiple view clustering \cite{MVC,MVC_CCA,MVC_co_reg} aims to enhance clustering performance by integrating different views.
Moreover, as the information explodes, we can get information from multiple sources.
Each source can contain multiple views that are fully aligned and available for clustering.
Combining data from multiple sources, multiple views may help us get better clustering performance.

However, several difficulties prevent us from combining different sources and views.
First, the views may be very heterogeneous.
Different views may have different feature spaces and distributions.
Second, the instances mapping between different sources may be incomplete and partially observed.
Different sources may have different instance sets, which means the instance mapping between different sources is not fully mapped.
Also in real-world problems, the instance mapping is often partially observed.
We may only get part of the instance mapping between different sources.
Third, the views in one source are generally more cohesive than the views across different sources.
This is very different from the traditional multi-view clustering problem.

\begin{figure}[t]
	\centering
	\includegraphics[width=0.9\columnwidth]{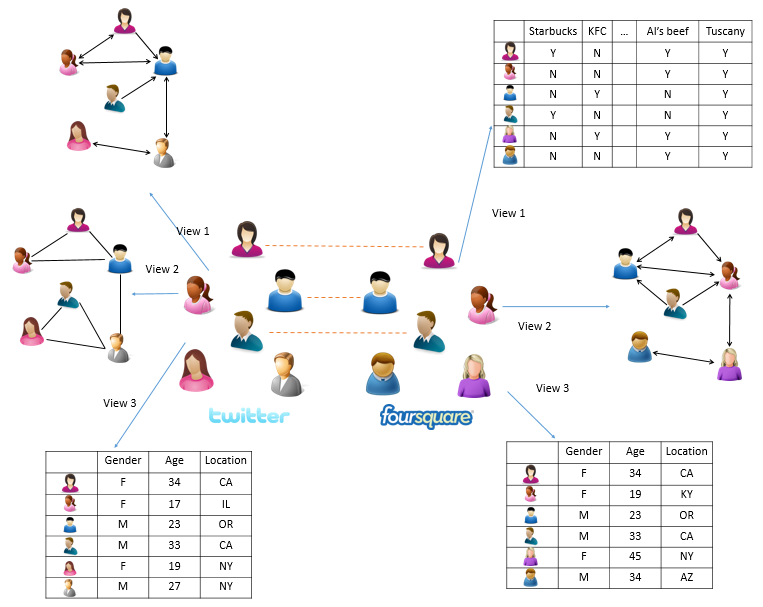}
	\caption{\small An Example of Social Networks.
		Twitter and Foursquare are two sources. Twitter contains user connection view, user interaction view and user profile view,
		while Foursquare contains user connection view, user check-in location view, user profile view.
		The user mapping between Twitter and Foursquare is incomplete and partially observed.
		}
	\label{sns}
\end{figure}

A good example is the social networks shown in Fig.~\ref{sns}.
People usually use several social network services simultaneously, \textit{e.g.}, those provided by Twitter and Foursquare. 
Each social network is an independent source containing several views that describe different aspects of the social network.
We can use the profile information of users for both Twitter and Foursquare (view 3 in Fig.~\ref{sns}),
the social connections of users (view 2 in Fig.~\ref{sns}), location check-in history, etc.
Views in a single source describe the characters of the same set of users and focus on different aspects.
The views in Twitter focus on the social activity aspects,  while the views in Foursquare focus more on the location based aspects.
Since not all people use both Twitter and Foursquare,
the user mapping between Twitter and Foursquare is incomplete and not one-to-one.
Furthermore, not all the shared users link their Twitter accounts with their Foursquare accounts.
We can only observe part of the mapping between Twitter and Foursquare.

Multi-view clustering \cite{MVC,LongYZ08,MVC_co_training} aims to utilize the multiple representations of instances in different features spaces to get better clustering performance.
However, most of the previous methods are based on the assumption that all the views are fully mapped/aligned.
They cannot deal with the multi-source multi-view scenario with incomplete/partial mapping across sources.
Although there are some previous studies on dealing with multiple incomplete view clustering \cite{ShaoMVC,PVC,flexibleMVC,PAKDDShao},
none of them are suitable for multi-source multi-view scenario.
They either cannot extend to more than two views or they do not treat the views in one source as a cohesive set.
All of the previous methods only use the known mapping information.
Furthermore, none of them try to extend the known mapping information to help improve clustering.
These challenges and emerging applications call for
novel clustering methods which can deal with multiple sources multiple views situations.
In this paper, we propose MMC (\textbf{M}ulti-source \textbf{M}ulti-view \textbf{C}lustering)
to integrate multiple sources for better clustering results.
The main contributions of this paper are summarized as follows:
\begin{enumerate}
	\setlength\itemsep{0em}
	\item This paper is the first one to investigate the multi-source multi-view clustering problem,
	where multiple sources containing multiple views are available for clustering and
	the instance mappings between sources are incomplete and partially known.
	\item We develop a method MMC,
	based on collective spectral clustering with discrepancy penalty within and between sources.
	MMC appreciates the cohesiveness of all the views in each source
	by pushing the latent feature matrice of the views in one source to a consensus for each source.
	MMC also considers the cross-source discrepancy by minimize the difference among the consensus latent feature matrices of different sources.
	\item The proposed MMC not only generates clusters from multiple sources
	but also tries to infer the unknown instance similarity mapping between sources to improve the clustering performance.
	By using the learned consensus latent feature matrices of different sources
	and the incomplete/partial mappings across sources,
	MMC generates the instance similarities across different sources,
	which will in turn help enhance the clustering performance.
	\item The proposed MMC is not limited to the multiple sources multiple views problem.
	In real-world applications, we may have multiple partial aligned views, \textit{i.e.,} views with different numbers of instances. We can group the views into groups where each group has the same set of instances. Thus each group can be viewed as a source and our method can be applied.
\end{enumerate}

The experiment results on three groups of real data show the effectiveness of the proposed method by comparing it with other state-of-art methods.

\section{Problem Formulation}
In this section, we will first define the problem of multi-source multi-view clustering.
Then we will start from the single source problem to
develop the objective function for multi-source multi-view clustering problem.
\subsection{Problem Definition}
Before we define the problem of multi-source multi-view clustering,
we summarize the notations in Table \ref{table_symbol}.
Let $\mathcal{S} = \{S^k\}_{k=1}^K$ denote the set of the $K$ available sources.
For each $S^k$, we have $S^k = \{X_i^k\}_{i=1}^{v_k}$,
where $X_i^k$ denotes the $i$-th view in source $k$ and
$v_k$ denotes the number of views in source $k$.
We assume that source k $(1\le k\le K)$ contains $n_k$ instances.
Let $ \mathcal{M} = \{M^{(i,j)}\}_{1\le i,j\le K}$
be the known instance mappings between sources,
where $M^{(i,j)} \in \mathbb{R}_{+}^{n_i\times n_j}$
denotes the instance mapping between sources i and j, and its element is defined by
\begin{equation}
M_{a,b}^{(i,j)} = \begin{cases}
1&\text{instance $ a $ in source $ i $ is mapped to } \\
& \text{instance $ b $ in source $ j $}.\\
0&\text{otherwise.}
\end{cases}
\end{equation}
\newcommand{\tabincell}[2]{\begin{tabular}{@{}#1@{}}#2\end{tabular}}
\begin{table}
	\caption{\small Summary of symbols and their meanings}
	\centering
	\begin{adjustbox}{max width=0.95\columnwidth}
		\begin{tabular}{|c|c|}
			\hline
			Symbols & Description\\
			\hline
			K & Number of sources.\\
			$v_k$ & Number of views for source $k$.\\
			$n_k$ & Number of instances in source $k$.\\
			$c_k$ & Number of clusters for source $k$.\\
			$X^{k}_i$ & i-th view in source $ k $\\
			$K^k_i$ & Kernel matrix for the $i$-th view in source $ k $\\
			$K_{U}$ & Kernel matrix of the matrix $ U $\\
			$L^k_i=(D^k_i)^{-1/2}K^k_i(D^k_i)^{-1/2}$ &
			\tabincell{l}{Normalized graph Laplacian for the i-th view in \\source $k$,
				where $D^k_i$ is a diagonal matrix consisting \\of the row sums of $K_i^k$.}\\
			$M^{(i,j)}\in \mathbb{R}^{n_i\times n_j}$ & Instance mapping between source $ i $ and $ j $.\\
			$W^{(i,j)}\in \mathbb{R}^{n_i\times n_j}$ & Indicator matrix between source $ i $ and $ j $\\
			$\alpha^k_i$ & Importance of view $ i $ in source $ k $\\
			$\beta^{(i,j)}$ & Importance of penalty between source $ i $ and $ j $\\
			$U^k_i\in \mathbb{R}^{n_k\times c_k}$& Latent feature matrix for view $ i $ in source $ k $\\
			$U^{k*}\in \mathbb{R}^{n_k\times c_k}$ & Consensus latent feature matrix for source $ k $\\
			\hline
		\end{tabular}
	\end{adjustbox}
	\label{table_symbol}
\end{table}

Our goal is to cluster the instances into $c_k$ clusters for each source k,
while considering the other sources by using the cross-source mapping in $\mathcal{M}$.
\subsection{Single Source Multiple Views Clustering}
Clustering with multiple views within a single source can be seen as traditional multi-view clustering problem.
However, in order to incorporate the cross-source disagreement,
we would like to get a consensus clustering solution for each source.
Let $K^k_i$ be the positive semi-definite similarity matrix or kernel matrix for view $ i $ in source $k$.
The corresponding normalized graph Laplacian will be $L^k_i = {D^k_i}^{-1/2}K^k_i{D^k_i}^{-1/2}$,
where $D^k_i$ is a diagonal matrix with the diagonal elements be the row sums of $K^k_i$.
To perform spectral clustering for a single view $ i $ in source $ k $,
as shown in \cite{MVC_co_reg,SC_tutorial,Andrew_sc},
we only need to solve the following
optimization problem for the normalized graph Laplacian $L^k_i$:
\begin{equation}
\max_{U^k_i\in \mathbb{R}^{n_k\times c_k}}~ tr((U^k_i)^TL^k_iU^k_i), ~~s.t. (U^k_i)^TU^k_i = \mathit{I},
\label{single_view}
\end{equation}
where $tr$ denotes the matrix trace,
and $ U^k_i $ can be seen as a latent feature matrix of view $i$ in source $ k $,
which can be given to the k-means algorithm to obtain cluster memberships.
For clustering multiple views in the same source,
we have to consider the disagreement between different views.
We enforce the learned latent feature matrix for each view to look similar
by regularizing them towards a common consensus.

Similar to the regularization in \cite{MVC_co_reg},
we define the discrepancy/dissimilarity between two latent feature matrices as:
\begin{equation}
D(U^k_i, U^k_j) = \|K_{U^k_i}-K_{U^k_j}\|_F^2,
\label{eq._in_source_disagree}
\end{equation}
where $K_{U^k_i}$ and $K_{U^k_j}$ are the similarity/kernel matrices for $U^k_i$ and $U^k_j$,
and $\|.\|_F$ denotes the Frobenius norm of the matrix.
We use linear kernel as the similarity measure in Eq.~(\ref{eq._in_source_disagree}).
Thus, we get $K_{U^k_i} = U^k_i(U^k_i)^T$ and $K_{U^k_j} = U^k_j(U^k_j)^T$.
Using the properties of trace and the fact that $(U^k_i)^TU^k_i=\mathit{I}$, $(U^k_j)^TU^k_j=\mathit{I}$,
Eq.~(\ref{eq._in_source_disagree}) can be rewritten as follows:
\begin{equation}
\begin{adjustbox}{max width=1\columnwidth}
$\displaystyle
\begin{split}
&D(U^k_i, U^k_j) = \|U^k_i(U^k_i)^T-U^k_j(U^k_j)^T\|_F^2 \\
&=tr\left(U^k_i(U^k_i)^T + U^k_j(U^k_j)^T - 2U^k_i(U^k_i)^TU^k_j(U^k_j)^T\right)\\
&=2c_k - 2tr(U^k_i(U^k_i)^TU^k_j(U^k_j)^T).
\end{split}
$
\end{adjustbox}
\end{equation}
Ignoring the constant terms,
we can get the discrepancy between two different latent feature matrices:
\begin{equation}
D(U^k_i, U^k_j) = -tr(U^k_i(U^k_i)^TU^k_j(U^k_j)^T).
\end{equation}
Then $ D(U^k_i, U^{k*}) = -tr(U^k_i(U^k_i)^TU^{k*}(U^{k*})^T)$
is the discrepancy between the $ i $-th view and the consensus.
Considering the discrepancy/dissimilarity between each view and the consensus in the same source,
the objective function for clustering all the views within the $ k $-th source will be as follows:
\begin{equation}
\small
\centering
\begin{adjustbox}{max width=0.9\columnwidth}
$\displaystyle
\begin{split}
\max_{\{U^k_{i}\},{U^{k*}}} \mathcal{J}_k &= \sum_{i=1}^{v_k}\left( tr({U^k_{i}}^TL^k_{i}U^k_{i}) + \alpha^k_{i}tr(U^k_{i}{U^k_{i}}^TU^{k*}{U^{k*}}^T)\right) ,\\
&\textit{s.t.}~~ {U^k_{i}}^TU^k_{i} = \mathit{I}, \forall 1\le i\le v_k, {U^{k*}}^TU^{k*} = \mathit{I}
\end{split}
$
\end{adjustbox}
\label{eq:multi}
\end{equation}
where $\mathcal{J}_k$ is the objective function for source $ k $,
$U^k_i$ is the latent feature matrix for the $ i $-th view in source $ k $,
$ U^{k*} $ is the consensus latent feature matrix for source $ k $,
and $\alpha^k_i$ is the relative importance of view $ i $ in source $ k $.
\subsection{Multiple Sources Multiple Views Clustering}
In this section, we will derive the objective function for Multi-source Multi-view Clustering problem.
We model it as a joint matrix optimization problem.

In order to incorporate multiple sources,
we need to add penalty between sources to the single source multiple views clustering objective function.
To appreciate the cohesiveness of the views within one source,
we learn the consensus latent feature matrix for each source and
only penalize the discrepancy between those consensus latent feature matrices.

Similar to the discrepancy function across views within one source,
the penalty function across sources should consider the discrepancy between
the consensus clustering results for the sources.
Since the mappings between sources are incomplete and partially known,
we cannot directly apply the same penalty function as in the single source multiple views clustering objective function.
However, by using the mapping matrices, we can project the learned latent feature matrix from one source to other sources,
${M^{(i,j)}}^TU^{i*}$ can be seen as projection of the instance in source $ i $ to the instances in source $ j $.
The penalty function for discrepancy between source $ i $ and source $ j $ is as follow:
\begin{equation}
\begin{adjustbox}{max width=0.85\columnwidth}
$\displaystyle
\widetilde{D}(U^{i*}, U^{j*}) = \|{M^{(i,j)}}^TU^{i*}({M^{(i,j)}}^TU^{i*})^T-U^{j*}{U^{j*}}^T\|_F^2.
$
\end{adjustbox}
\label{eq:penalty}
\end{equation}

Observing that the known mapping between two sources is one-to-one
and it is reasonable to assume that the unknown part is one-to-at-most-one, which means one instance in one source can be mapped to at most one instance in the other source.
So we can approximately assume that $ M^{(i,j)}M^{(i,j)T} = I $ to help simplify the penalty function.
The Eq.~{\ref{eq:penalty}} can be expressed as:
\begin{equation}
\begin{adjustbox}{max width=0.872\columnwidth}
$\displaystyle
tr\left(\left({M^{(i,j)}}^TU^{i*}U^{i*T}M^{(i,j)} -U^{j*}U^{j*T}\right) \left({M^{(i,j)}}^TU^{i*}U^{i*T}M^{(i,j)} -U^{j*}U^{j*T}\right)\right)$
\end{adjustbox}
\end{equation}
Using the fact that $ U^{i*T}U^{i*} = I $, $ tr(U^{i*T}U^{i*}) = c_i $, and $ M^{(i,j)}M^{(i,j)T} = I $,
and ignoring the constant terms, the penalty function for discrepancy is
\begin{equation}
\small
\widetilde{D}(U^{i*}, U^{j*}) = -tr\left(U^{j*}U^{j*T}{M^{(i,j)}}^TU^{i*}U^{i*T}M^{(i,j)}\right)
\end{equation}
Adding the penalty function between sources to the single source multiple views clustering
objective function for all the sources, we get
\begingroup\makeatletter\def\f@size{7.5}\check@mathfonts
\begin{equation}
\begin{split}
\label{final}
\max_{U^k_{i},U^{k*}(1\le k\le K, 1\le i\le v_k)} \mathcal{O} = \sum_{k=1}^{K} \mathcal{J}_k - \sum_{i\ne j, 1\le i,j\le K}\beta^{(i,j)}\widetilde{D}(U^{i*}, U^{j*}),
\end{split}
\notag
\end{equation}
\endgroup
where $ \beta^{(i,j)} \ge 0$ is a parameter controlling the balance between the objective function for individual source and the inconsistency across sources.

\section{Optimization and MMC framework}
The proposed MMC framework simultaneously optimizes the latent featue matrices
in multiple sources and infers the cross-source instance similarity mappings to help enhance the clustering performance.
To optimize the objective function in Eq.~(\ref{final}),
we employ an alternating scheme, that is,
we optimize the objective function with respect to one variable while fixing others.
Basically, we optimize the objective function using two stages.
First, maximizing $\mathcal{O}$ over $U^k_{i}$s with fixed $U^{k*}$s.
Second, maximizing $\mathcal{O}$ over $U^{k*}$s with fixed $U^k_{i}$s.
We repeat these two steps, until it converges.

\begin{figure}
	\centering
	\includegraphics[width=0.85\columnwidth]{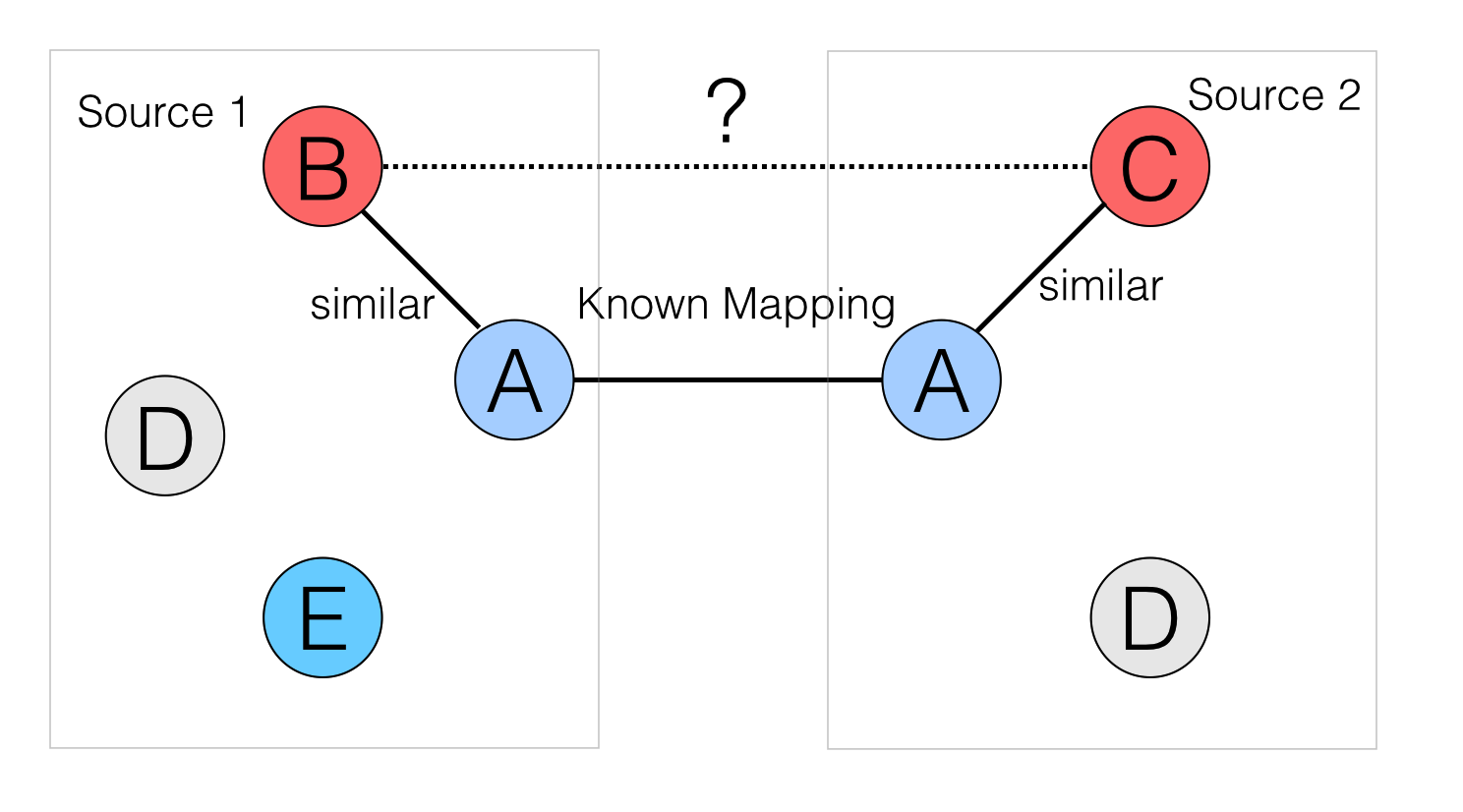}
	\caption{\small Principle of Transitivity on Similarity.}
	\label{fig_similar}
\end{figure}

Solving the optimization, we can iteratively learn the latent feature matrices for each source.
However,
when only a small portion of instances mapping between sources are observed,
the clustering performance is affected by the incompleteness of the instance mapping across sources.
Inferring the exact instance mapping is really challenging and usually additional information is required to infer such anchor link \cite{anchorlink,Zhang:2014:MBM:2623330.2623645}.
However, instead of inferring the exact instance mapping,
we can try to infer the similarity mapping.
This idea is based on the \textbf{Principle of Transitivity on Similarity}:
\begin{theorem}Instances similar to the same instance in different sources should be similar.\end{theorem}
Fig.~\ref{fig_similar} illustrates the principle.
In Fig.~\ref{fig_similar}, B and C are both similar to  A in different sources.
Given the known mapping between A in the two sources, we can infer that B and C are similar.
MMC tries to use the similarity transitivity principle to help infer the cross-source instance similarity and to help improve the clustering performance.
Next, we will talk about each step in the MMC framework in detail.
\subsection{Initialization}
Since the efficiency of the iterative optimization is affected by the initialization step,
in this paper, we learn the initial value of $ U_i^k $ and $ U^{k*} $ rather than random initialization.
For each $ U_i^k $, we solve Eq.~(\ref{single_view}) to get an initial value.
As we described in the previous section, Eq.~(\ref{single_view}) is just the objective function for single view clustering, without considering the relation among views and sources.
The solution $ U^k_i $ is given by the top-$ c_k $ eigenvectors of the Laplacian $ L_i^k $.

For each $ U^{k*} $, we just solve Eq.~(\ref{eq:multi}) to get the initialization value with the initial $ U^k_i $.
The objective function can be written as:
\begin{equation}
\small
\centering
\begin{split}
\max_{U^{k*}} ~& tr\left({U^{k*}}^T \left(\sum_{i=1}^{v_k}\alpha^k_{i} U^k_{i}{U^k_{i}}^T \right)U^{k*} \right),\\
&\textit{s.t.}~~ {U^{k*}}^TU^{k*} = \mathit{I}
\end{split}
\label{eq:init}
\end{equation}
The solution is given by the top-$ c_k $ eigenvectors of the modified Laplacian $ \sum_{i=1}^{v_k}\alpha^k_{i} U^k_{i}{U^k_{i}}^T$.

In the previous section, we assume that the mapping matrix between two sources is semi-orthogonal,
\textit{i.e.}, $ M^{(i,j)}M^{(i,j)T} = I $.
However, we can only get part of the mapping information due to the incompleteness and partial known property of real-world problem.
The mapping matrix between two sources can be expressed in two parts $ M_0^{(i,j)} $ and $ M_1^{i,j} $:
\begin{equation}
M^{(i,j)} = \begin{bmatrix}
M_0^{(i,j)} & 0 \\
0 & M_1^{(i,j)}
\end{bmatrix},
\end{equation}
where $ M_0^{(i,j)} $ represents the known mapping between source $ i $ and $ j $
and $ M_1^{(i,j)}$ represents the unknown part.

It is easy to find that $ M_0^{(i,j)}M_0^{(i,j)T} = I$.
We only need to initialize the unknown part to make it semi-orthogonal.
One natural way to estimate the unknown mapping is to use the instance similarity among the instances between two sources.
Using the similarity transitivity principle, the instance similarity between two sources $ i $ and $ j $ can be estimated by:
$\small
K_{U^{i*}}\begin{bmatrix}
M_0^{(i,j)} & 0 \\
0 & 0
\end{bmatrix}
K_{U^{j*}},
$
where $ K_{U^{i*}} $ and $ K_{U^{j*}} $ are the two kernel matrices for latent feature $ U^{i*} $ and $ U^{j*} $.
It is worth to note that the kernel matrices $K_{U^{i*}}$ and $K_{U^{j*}}$ can be seen as the similarity matrices of the latent features for sources $ i $ and $ j $.
The similarity transitivity principle allows us to use the known instance mapping as a bridge to connects unmapped instances in two sources.
Using Fig.~\ref{fig_similar} as an example,
$\left(K_{U^{i*}}\right)_{b,a}$ provides the similarity between instances B and A in source $ i $, while $\left(K_{U^{j*}}\right)_{a,c}$ provides the similarity between instances A and C in source $ j $.
If both instances A in both sources $ i $ and $j$ are
mapped through $ M^{(i, j)} $,
$\left(K_{U^{i*}}\begin{bmatrix}
M_0^{(i,j)} & 0 \\
0 & 0
\end{bmatrix}K_{U^{j*}}\right)_{b,c}$, denoted by $ (\widetilde{M}^{(i,j)})_{b,c} $,
will provide the estimated similarity between instance B in source $ i $ and instance C in source $ j $.
So for the unmapped instances, we can have an estimated similarity mapping $ \widetilde{M}_1^{(i,j)} $.
Then we can orthogonalize it using SVD or other orthogonalization methods.

\subsection{Maximizing $\mathcal{O}$ over $U^k_{i}$s with fixed $U^{k*}$s}
With fixed $U^{k*}$s, for each $U^k_{i}$ we only need to maximize part of $\mathcal{J}_k$.
\begin{equation}
\small
\begin{split}
\max_{U^k_{i}}~ \mathcal{L} &=tr({U^k_{i}}^TL^k_{i}{U^k_{i}})+\alpha^k_{i}tr({U^k_{i}}{U^k_{i}}^TU^{k*}{U^{k*}}^T)\\
&=tr({U^k_{i}}^T(L^k_{i}+\alpha^k_{i}U^{k*}{U^{k*}}^T){U^k_{i}})\\
&\textit{s.t.}~~ {U^k_{i}}^TU^k_{i} = \mathit{I}
\end{split}
\label{eq:op1}
\end{equation}
This is a standard spectral clustering objective on source $k$ view $i$
with modified graph Laplacian $L^k_{i}+\alpha^k_{i}U^{k*}{U^{k*}}^T$.
According to \cite{Andrew_sc},
the solution $U^k_{i}$ is given by the top-$c_k$ eigenvectors of this modified Laplacian.
With fixed $U^{k*}$s, we can calculate each $U^k_{i}$ to maximize the objective function.\\

\subsection{Maximizing $\mathcal{O}$ over $U^{k*}$s with fixed $U^k_{i}$s}
With fixed $U^k_{i}$s, for each $U^{k*}$, we only need to maximize:
\begin{equation}
\resizebox{\hsize}{!}{$\displaystyle
	\begin{split}
	\max_{U^{k*}} \mathcal{Q} &= \sum_{i=1}^{v_k}\alpha^k_{i}tr(U^k_{i}{U^k_{i}}^TU^{k*}{U^{k*}}^T) - \sum_{1\le j\le K,j\ne k} \beta^{(k,j)}\widetilde{D}(U^{k*}, U^{j*})\\
	&=\sum_{i=1}^{v_k}\alpha^k_{i}tr(U^k_{i}{U^k_{i}}^TU^{k*}{U^{k*}}^T) \\
	&+ \sum_{j\ne k} \beta^{(k,j)} tr\left(U^{j*}U^{j*T}{M^{(k,j)}}^TU^{k*}U^{k*T}M^{(k,j)}\right)\\
	&= tr\left( U^{k*T}L^{k*}U^{k*}\right)\\
	&\textit{s.t.}~~ {U^{k*}}^TU^{k*} = \mathit{I}.\\
	\end{split}$
}
\label{eq:update_star}
\end{equation}
where $$ L^{k*} = \sum_{i=1}^{v_k}\alpha^k_{i}U^k_{i}{U^k_{i}}^T+\sum_{j\ne k}\beta^{(k,j)}M^{(k,j)} U^{j*}U^{j*T}{M^{(k,j)}}^T$$
The solution $U^{k*}$ is given by the top-$c_k$ eigenvectors of this modified Laplacian $L^{k*}$.
Thus, with fixed $U^{k}_i$s, we can calculate the consensus $U^{k*}$ for each of the sources to maximize the objective function.

\subsection{Infer the Similarity Mapping between Sources}
Using above optimization method, we can iteratively learn the latent feature matrices for each source.
However, when the number of partially observed mapping is limited,
\textit{i.e.}, when only a small number of instances mapping between sources are observed,
the estimated initial similarity mapping between two sources may not be accurate.
Hence the improvement of clustering performance is limited.

Based on the similarity transitivity principle, MMC proposes to use the learned latent feature matrices for multiple sources
to help infer the similarity mapping across sources.

Similar to the initialization, we use the instance mapping between two sources as a bridge to help transfer the similarity.
The new estimated instance similarities between two sources can be written as:
\begin{equation}
\widetilde{M}^{(i,j)} = K_{U^{i*}}M^{(i,j)}K_{U^{j*}},
\end{equation}
where $ K_{U^{i*}} $ and $ K_{U^{j*}} $ are the two kernel matrices for latent feature matrices $ U^{i*} $ and $ U^{j*} $.
In our experiment, we use linear kernel for the latent feature matrices.

This estimated similarity mapping includes every instance across two sources.
However, we want to preserve the already known instance mapping and
only update the instance similarity mapping for instances whose mappings are unknown.
We introduce the indicator matrix $W^{i,j}$, which has the same dimension as $M^{(i,j)}$ and was initialized with only 0 and 1.
$W^{(i,j)}_{ab}$ equals to 1 if the mapping between $ a $-th instance from source $ i $
and the $ b $-th instance from source $ j $ is known, and 0 if unknown.
The similarity mapping between source $ i $ and source $ j $ and is updated as follows:
\begin{align}
\begin{adjustbox}{max width=0.85\columnwidth} $\displaystyle
M^{(i,j)} \leftarrow W^{(i,j)}\circ M^{(i,j)} + ( \mathbf{1} -W^{(i,j)})\circ \widetilde{M}^{(i,j)},
$
\end{adjustbox}
\label{update_mapping}
\end{align}
where $\circ$ indicates the element-wise multiplication,
$\widetilde{M}^{(i,j)} = U^{i*}{U^{i*}}^TM^{(i,j)}U^{j*}{U^{j*}}^T$
and $ \mathbf{1} $ is an all-one matrix.
By using the indicator matrix $W^{(i,j)}$ and element-wise multiplication,
we can only update the unknown part of the mapping,
and preserve the known part.
Once we have a better mapping across sources, it will help learn better latent feature matrices.
The better latent feature matrices will in-turn help infer the similarity mapping.
This iteration continues until it converges.

\subsection{MMC framework}
The algorithm for the MMC framework is shown as Algorithm \ref{algorithm_MMC}.
We first calculate the kernel matrices and
the corresponding normalized graph Laplacian matrices for every view.
In all the experiments throughout the paper, 
we use Gaussian kernel for computing the similarities unless mentioned otherwise. 
The standard deviation of the kernel is set equal to 
the median of the pair-wise Euclidean distances between the data points.
We then initialize the latent feature matrices $ \{U_i^k\} $, $ \{U^{k*} \} $ and the instance mappings $ M^{(i,j)} $.
Then we iteratively update $U_i^k$s, $U^{k*}$s and $M^{(i,j)}$s until they all converge.

\begin{algorithm}                      
	\caption{\small MMC framework}          
	\label{algorithm_MMC}                           
	\begin{algorithmic}[1]                  
		\floatname{algorithm}{Procedure}
		\renewcommand{\algorithmicrequire}{\textbf{Input:}}
		\renewcommand{\algorithmicensure}{\textbf{Output:}}
		\REQUIRE Data matrices for every view from each source $\{X^k_i\}$. 
		Instance mappings between sources $\{M^{(i,j)}\}$.
		Indicator matrices $\{W^{(i,j)}\}$.
		The number of clusters for each source $\{c_k\}$. 
		Parameters $ \{\alpha^k_i\} $ and $\{\beta^{(i,j)}\}$.
		\ENSURE Clustering results for each source.
		\STATE Calculate $ K^k_i $ and $L^k_i$ for every $ k $ and $ i $.
		\STATE Initialize $U_i^k$  for every $ k $ and $ i $.
		\STATE Initialize $U^{k*}$ by solving Eq.(\ref{eq:init}).
		\REPEAT 
		\REPEAT
		\STATE Update each $U_i^k$ by solving Eq.~(\ref{eq:op1}).
		\STATE Update each $U^{k*}$ by solving Eq.~(\ref{eq:update_star}).
		\UNTIL{objective function $ \mathcal{O} $ converges.}
		\STATE Update the similarity mappings using Eq.~(\ref{update_mapping}).
		\UNTIL{mappings between sources converge.}
		\STATE Apply k-means on $U^{k*}$ for every source $ k $.
	\end{algorithmic}
\end{algorithm}

\section{Experiments and Results}
In this section, we compare  MMC framework with a number of baselines on three real-world data sets.
\subsection{Comparison Methods}
We compare the proposed MMC method with several state-of-art methods.
Since no previous methods can be directly applied to the multi-source multi-view situation,
in order to compare with the previous methods, we make some changes.
The details of comparison methods is as follows:
\begin{itemize}
	\item \textbf{MMC:} MMC is the clustering framework proposed in this paper,
	which applies collective spectral clustering with discrepancy penalty across sources.
	The parameter $\alpha$ is set to 0.1 and $\beta$ is set to 1 for all the views and sources throughout the experiment.
	\item \textbf{Concat:} Feature concatenation is one way to integrate all the views.
	We concatenate views within each source, so each source is a concatenated view.
	Since
	the instances between sources are not fully aligned,
	we extend each source by adding pseudo instances (average features).
	Thus, sources are fully aligned after extension.
	We then apply PCA and k-means to get the clustering results.
	\item \textbf{Sym-NMF} Symmetric non-negative matrix factorization is proposed in
	\cite{sym_nmf} as a general framework for clustering.
	It factorizes a symmetric matrix containing pairwise similarity values.
	To apply Sym-NMF to multi-source multi-view situation,
	we apply Sym-NMF to every view from each source to get the latent feature matrices.
	Then we combine all the latent feature matrices in the same source to produce the final clustering results.
	\item \textbf{MultiNMF:} MultiNMF is one of the state-of-art multi-view clustering methods
	based on joint nonnegative matrix factorization \cite{sdm2013_liu}.
	MultiNMF formulates a joint matrix factorization process with the constraint
	that pushes clustering solution of each view towards a common consensus instead of fixing it directly.
	Throughout the experiment, the parameter $\lambda_v$ is set to 0.01 as in the original paper.
	\item \textbf{CoReg:} CoReg is the centroid based multi-view clustering method proposed in \cite{MVC_co_reg}.
	It aims to get clusters that are consistent across views by co-regularizing the clustering hypotheses.
	Throughout the experiment, the parameter $\lambda_v$ is set to 0.01 as suggested in the original paper.
    \item \textbf{CGC:} CGC \cite{flexibleMVC} is the most recent work that
    deals with many-to-many instance relationship, which is similar to incomplete instance mapping.
    In order to run the CGC algorithm, we generated the relations between views within one source, which is complete one-to-one mapping.
    We also generate the relations between views across sources, which is incomplete and partially known.
    We run CGC on all the views across sources and report the best performance for each source.
    In the experiment, the parameter $\lambda$ is set to 1 as suggested in the original paper.
\end{itemize}

It is worth to note that the two multi-view clustering methods MultiNMF  and Co-Reg only work with views that are fully aligned.
In our experiments, only views from the same source are fully mapped/aligned.
Views across different sources are partially mapped.
We apply MultiNMF and Co-Reg in two ways.

\begin{figure}
	\centering
	\includegraphics[width=0.8\columnwidth]{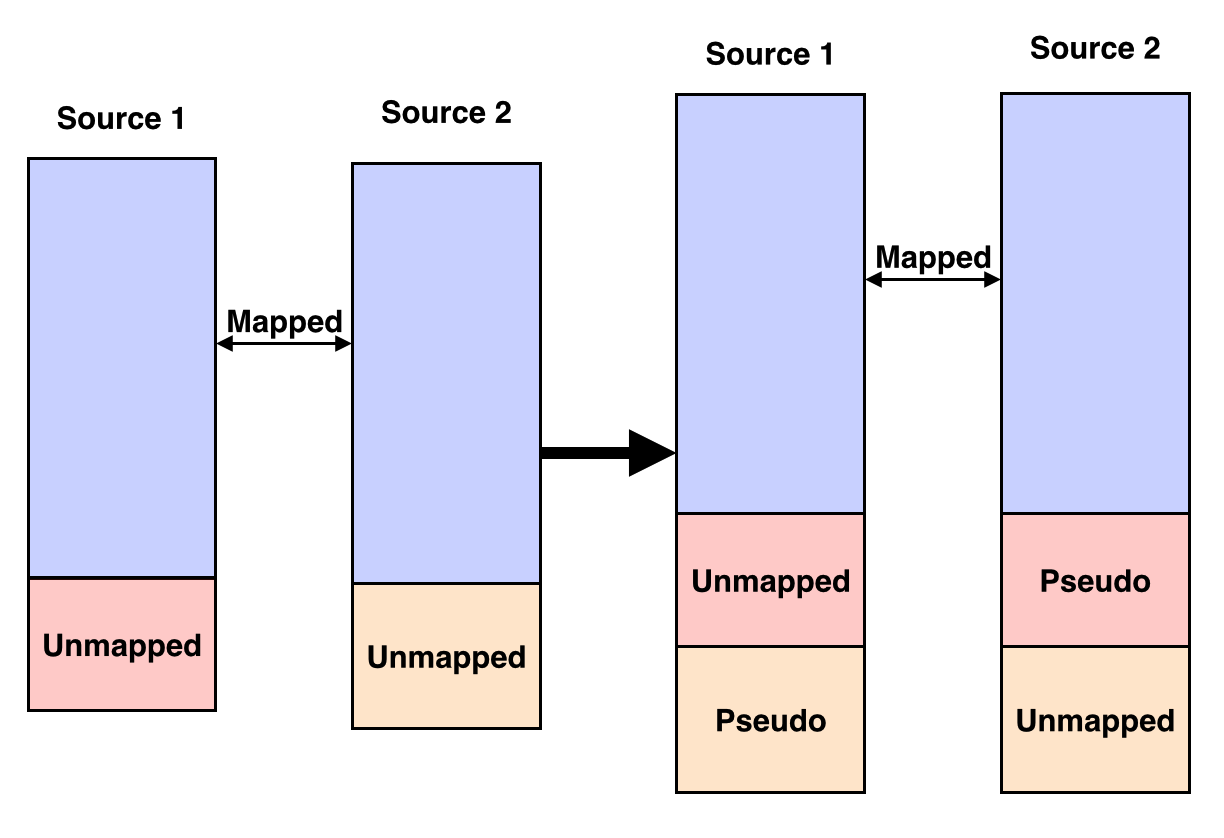}
	\caption{\small Source alignment}
	\label{fig:pseudo}
\end{figure}

The first way is to apply them to every single source, denoted as \textbf{MultiNMF-S} and \textbf{CoReg-S}.
Thus, both MultiNMF-S and Co-Reg-S only co-regularize the views within a source
without considering the discrepancy between sources.
The second way is to apply MultiNMF and Co-Reg to multiple sources, denoted as \textbf{MultiNMF-M} and \textbf{CoReg-M}.
However, the views across sources are not fully mapped/aligned.
In order to apply MultiNMF and Co-Reg,
we align the sources by adding average pseudo instances to every source.
As shown in Fig.~\ref{fig:pseudo}, for the unmapped instances in one source, 
we created the corresponding pseudo instances in other sources.
Thus, the instances from different sources are fully mapped.
MultiNMF and Co-Reg are then applied to all the aligned views, and performance is reported for every source.

In our experiments,
we use Gaussian kernel for computing the similarities unless mentioned otherwise.
The standard deviation of the kernel is set equal to
the median of the pair-wise Euclidean distances between the data points.
K-means is used to get clustering results for all the methods.
For each setting, we run k-means 20 times and report the average performance.

\subsection{Dataset}
In this paper, three groups of real-world data sets are used to evaluate the proposed MMC method.
The important statistics of them are summarized in Table \ref{tab:data_stat}.
\begin{table}
	\centering
	\caption{\small Statistics of the datasets}
	\label{tab:data_stat}
	\begin{adjustbox}{max width=0.9\columnwidth}
	\begin{tabular}{|c|c|c|c|}
		\hline
		\textbf{data} & \textbf{size} & \textbf{\# view} & \textbf{\# cluster}\\
		\hline
		\hline
		\textbf{Dutch} & 2000 & 6 & 10\\
		\hline
		\textbf{USPS} & 2000 & 2 & 10\\
		\hline
		\hline
		\textbf{English}& 1200  & 2 & 6\\
		\hline
		\textbf{Translation} & 1200 & 2 & 6\\
		\hline
		\hline
		\textbf{BBC} & 352 & 2 & 6\\
		\hline
		\textbf{Reuters} & 294 & 2 & 6\\
		\hline
		\textbf{Guardian} & 302 & 2 & 6\\
		\hline
	\end{tabular}
	\end{adjustbox}
\end{table}
\begin{itemize}
	\item \textbf{Dutch-USPS}
	This data set comes from two sources, UCI Handwritten Dutch digit numbers
	and USPS digit data.
	The first source, \textbf{Dutch}\footnote{https://archive.ics.uci.edu/ml/datasets/Multiple+Features},
	consists of 2000 examples of handwritten numbers '0'-'9' (200 examples per class) 
	extracted from a collection of Dutch utility maps.
	All the examples have been digitized in binary images. 
	Each example is represented in the following six views:
	(1) 76 Fourier coefficients of the character shapes, 
	(2) 216 profile correlations, 
	(3) 64 Karhunen-Love coefficients,
	(4) 240 pixel averages in $2\times 3$ windows,
	(5) 47 Zernike moments,
	and (6) 6 morphological features.
	The second source, \textbf{USPS}\footnote{http://www.cs.nyu.edu/~roweis/data.html}, consists of digit images with size $ 16\times 16$ for numbers `0'-`9'.
	We randomly select 2000 examples corresponding to the examples in first source.
	From USPS data, we extract two views, the original pixel feature with dimension of 256
	and the Gaussian similarity matrix between examples with dimension of 2000.
	\item \textbf{English-Translations:}
	This data contains two sources,
	the original Reuters news documents written in English,
	and the machine translations in other four languages (French, German, Spanish and Italian) in 6 topics \cite{Amini09learningfrom}.
	From the first source, \textbf{English}, we use the document-term matrix
	and the cosine similarity matrix of the documents as two views.
	From the second source, \textbf{Translation},
	we extract the document-term matrices from French and German as two views.
	We randomly sample 1200 documents from the first source in a balanced manner,
	with each category having 200 documents.
	We then select the corresponding 1200 documents from the second source.
	\item \textbf{News Text data\footnote{http://mlg.ucd.ie/datasets/3sources.html}:}
	This news data has three sources:
	BBC, Reuters, and The Guardian.
	In total there are 948 news articles covering 416 distinct news stories 
	from the period February to April 2009.
	Thus, the articles from these three sources are naturally partially mapped.
	Of these distinct stories,
	169 were reported in all three sources,
	194 in two sources, and 53 appeared in a single news source.
	Each story was annotated with one of the six topical labels: business, entertainment, health, politics, sport, technology.
	From each source, we extract two views, the document-term matrix and the cosine similarity matrix.
	From the three sources, we create three sets of data,
	\textbf{BBC}-\textbf{Reuters} (239 mapped instances), \textbf{BBC}-\textbf{Guardian} (250 mapped instances) and \textbf{Reuters}-\textbf{Guadian} (212 mapped instances).
\end{itemize}

It is worth noting that both Dutch-USPS and English-Translation data are one-to-one fully mapped. 
In our experiments, we randomly delete part of the mappings across different sources.

\subsection{Results}
The results for Dutch-USPS data and English-Translation data are shown in Table \ref{tab:two_set}.
The results are obtained under 60\% known mappings .
We report the NMI (Normalized Mutual Information) for each source in Table \ref{tab:two_set}.

From Table \ref{tab:two_set}, we can observe that
the proposed MMC framework outperforms all the other comparison methods on
both Dutch-USPS and English-Translation data.
For the Dutch-USPS data, although CoReg-M and CGC are close to MMC (less than 4\%) on Dutch,
MMC outperforms the other methods on USPS by at least 10 \%.
We can also observe that MultiNMF-M and CoReg-M perform better than MultiNMF-S and CoReg-S on Dutch-USPS.
However, the performance of the multi-source methods is worse than single-source method on English-Translation.
This suggests that combining multiple sources only using the incomplete instance mappings may even hurt the performance.
The proposed MMC methods, however, iteratively discovers the similarity among unmapped instances and uses the similarity to help learning.

We also reported the results on three sets of the news text data (BBC-Reuters, BBC-Guardian and Reuters-Guardian) in Table \ref{tab:result_3source}.
From Table \ref{tab:result_3source},
we can see that MMC outperforms other comparison methods by a large margin in most cases.
On Reuters-Guardian, although MultiNMF-M is slightly better than MMC on Guardian,
MMC is still better than all the other baselines on Reuters.

From Table \ref{tab:two_set} and Table \ref{tab:result_3source},
we can observe that MMC outperforms other comparison methods in most cases for all the three groups of data.
We can also conclude that MMC reduces the impact of negative transfer by iteratively learns the similarity among unmapped instances and takes advantage of it.
The other reason why MMC can have a better performance for all the sources is that  MMC treats the views within each source as a cohesive unit for clustering
while considering discrepancy/disagreements between sources.

\begin{table}[t]
	\centering
	\caption{\small  NMI for \textbf{Dutch}-\textbf{USPS} and \textbf{English}-\textbf{Translation} at 60\% cross source mapping known}
	\label{tab:two_set}
	\begin{adjustbox}{max width=0.95\columnwidth}
	\begin{tabular}{|c|c|c|c|c|}
		\hline
		Method& \textbf{Dutch} & \textbf{USPS} & \textbf{English} & \textbf{Translation}\\
		\hline
		\hline
		Concat	&	0.5734	&	0.3916	&	0.1914	&	0.1621	\\ \hline
		Sym-NMF	&	0.7778	&	0.3005	&	0.2783	&	0.1527	\\ \hline
		MultiNMF-S	&	0.5382	&	0.4010	&	0.3413	&	0.2708	\\ \hline
		MultiNMF-M	&	0.7585	&	0.4700	&	0.342	&	0.2164	\\ \hline
		CoReg-S	&	0.7503	&	0.4044	&	0.3381	&	0.2874	\\ \hline
		CoReg-M	&	0.7886	&	0.5257	&	0.2187	&	0.2198	\\ \hline
		CGC	&	0.7851	&	0.2780	&	0.2636	&	0.2536	\\ \hline
		MMC	&	\textbf{0.8248}	&	\textbf{0.6348}	&	\textbf{0.3528}	&	\textbf{0.3073}	\\ \hline
	\end{tabular}
	\end{adjustbox}
\end{table}
\begin{table}[t]
	\centering
	\caption{\small NMI for News Text Data}
	\label{tab:result_3source}
	\begin{adjustbox}{max width=0.95\columnwidth}
	\begin{tabular}{|c|c|c|c|c|c|c|}
		\hline
		\multirow{2}{*}{Method} & \multicolumn{2}{c|}{\textbf{BBC}-\textbf{Reuters}} & \multicolumn{2}{c|}{\textbf{BBC}-\textbf{Guardian}} & \multicolumn{2}{c|}{\textbf{Reuters}-\textbf{Guardian}}\\
		\hhline{~------}
		& BBC & Reuters & BBC & Guardian & Reuters & Guardian\\
		\hline
		Concat	&	0.3003	&	0.3118	&	0.3344	&	0.3603	&	0.2994	&	0.3073	\\ \hline
		Sym-NMF	&	0.4414	&	0.443	&	0.4261	&	0.4444	&	0.4332	&	0.4325	\\ \hline
		MultiNMF-S	&	0.4799	&	0.4158	&	0.4637	&	0.4677	&	0.4275	&	0.4656	\\ \hline
		MultiNMF-M	&	0.5453	&	0.5127	&	0.4539	&	0.5243	&	0.5372	&	\textbf{0.5465}	\\ \hline
		CoReg-S	&	0.5488	&	0.5273	&	0.5532	&	0.5393	&	0.5273	&	0.536	\\ \hline
		CoReg-M	&	0.4311	&	0.4615	&	0.4927	&	0.5103	&	0.4644	&	0.454	\\ \hline
		CGC	&	0.4378	&	0.4159	&	0.4354	&	0.4171	&	0.4338	&	0.3921	\\ \hline
		MMC	&	\textbf{0.5714}	&	\textbf{0.5874}	&	\textbf{0.5632}	&	\textbf{0.5903}	&	\textbf{0.5639}	&	0.5377	\\ \hline
	\end{tabular}
	\end{adjustbox}
\end{table}
\begin{table}[t]
\centering
\small
\caption{\small NMI on Dutch-USPS with different clusters number}
\label{tab:clust_1}
\begin{adjustbox}{max width=0.95\columnwidth}
	\begin{tabular}{|c|c|c|c|c|c|}
		\hline
		Method&$c_1 = 2$&$c_1 = 4$&$c_1 = 6$&$c_1 = 8$&$c_1 = 10$\\
		\hline
		\hline
		Concat(D)	&	0.7806	&	0.5847	&	0.5499	&	0.5737	&	0.5734	\\ \hline
		Sym-NMF(D)	&	0.9652	&	0.8907	&	0.8393	&	0.7912	&	0.7778	\\ \hline
		MultiNMF-S(D)	&	0.9041	&	0.6996	&	0.5687	&	0.5917	&	0.5382	\\ \hline
		MultiNMF-M(D)	&	0.9652	&	0.8717	&	0.8086	&	0.7460	&	0.7585	\\ \hline
		CoReg-S(D)	&	0.9387	&	0.8879	&	0.8322	&	0.7807	&	0.7503	\\ \hline
		CoReg-M(D)	&	0.9652	&	0.9035	&	0.8205	&	0.8295	&	0.7886	\\ \hline
		CGC(D)	&	0.9652	&	0.9108	&	0.8345	&	0.7976	&	0.7851	\\ \hline
		MMC(D)	&	\textbf{0.9652}	&	\textbf{0.9433}	&	\textbf{0.8897}	&	\textbf{0.8311}	&	\textbf{0.8248}	\\ \hline \hline
		Concat(U)	&	0.4877	&	0.3470	&	0.3338	&	0.3730	&	0.3916	\\ \hline
		Sym-NMF(U)	&	0.5708	&	0.4412	&	0.3465	&	0.2875	&	0.3005	\\ \hline
		MultiNMF-S(U)	&	0.6261	&	0.6070	&	0.4783	&	0.4306	&	0.4010	\\ \hline
		MultiNMF-M(U)	&	0.7711	&	0.5006	&	0.6147	&	0.5585	&	0.4700	\\ \hline
		CoReg-S(U)	&	0.5375	&	0.5116	&	0.4207	&	0.3778	&	0.4044	\\ \hline
		CoReg-M(U)	&	0.5676	&	\textbf{0.7604}	&	0.5023	&	0.5146	&	0.5257	\\ \hline
		CGC(U)	&	0.5708	&	0.4513	&	0.2866	&	0.2642	&	0.2780	\\ \hline
		MMC(U)	&	\textbf{0.7753}	&	0.7435	&	\textbf{0.7109}	&	\textbf{0.6856}	&	\textbf{0.6348}	\\ \hline
	\end{tabular}
\end{adjustbox}
\end{table}

\subsection{Parameter Study}
\begin{figure}[ht]
	\centering
	\begin{minipage}[b]{0.80\columnwidth}
		\centering
		\includegraphics[width=\textwidth]{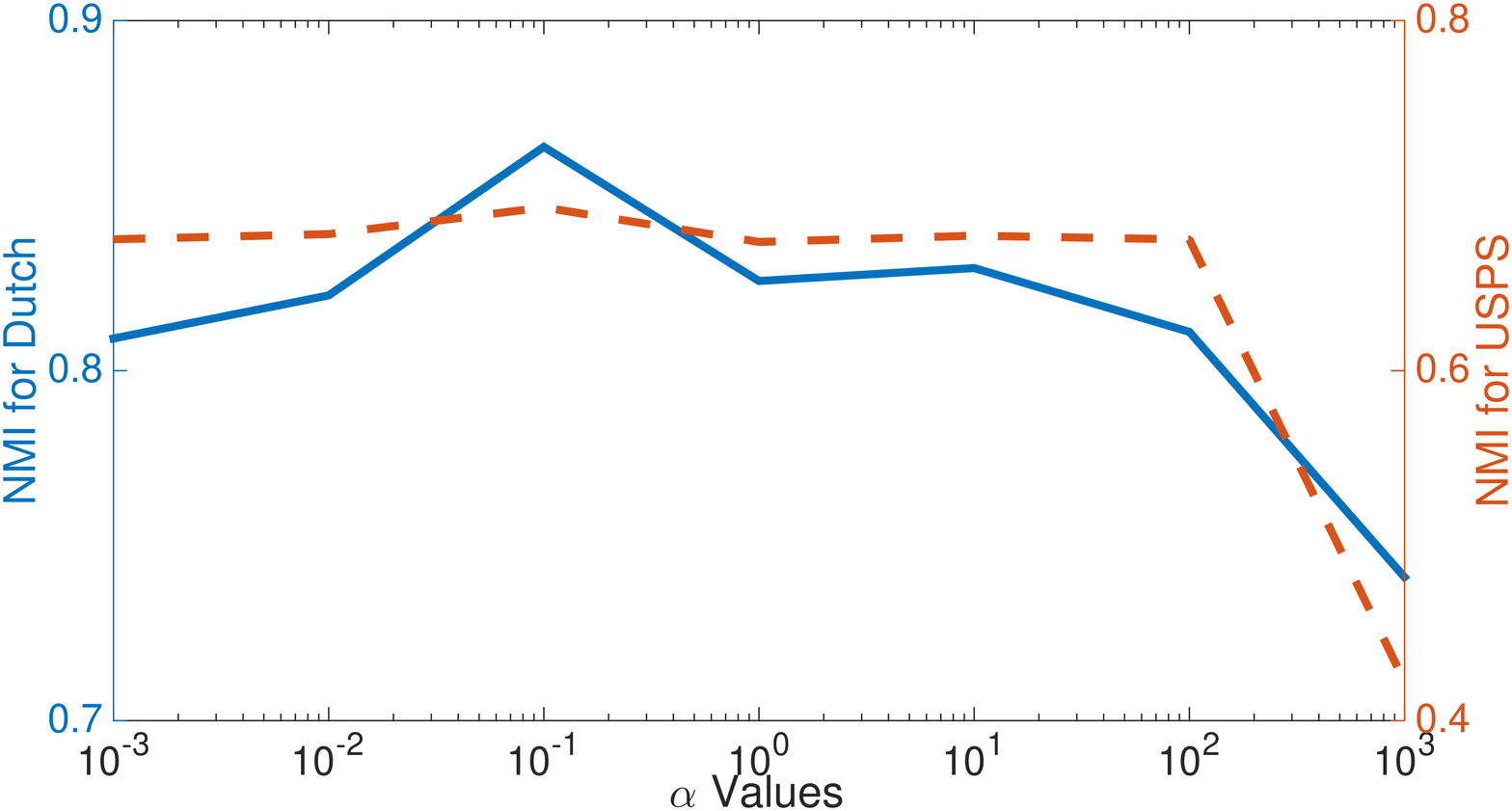}
		\caption{\small NMI v.s. $ \alpha_i^k $ on Dutch-USPS.}
		\label{fig:alpha}
	\end{minipage}
	\\
	\begin{minipage}[b]{0.80\columnwidth}
		\centering
		\includegraphics[width=\textwidth]{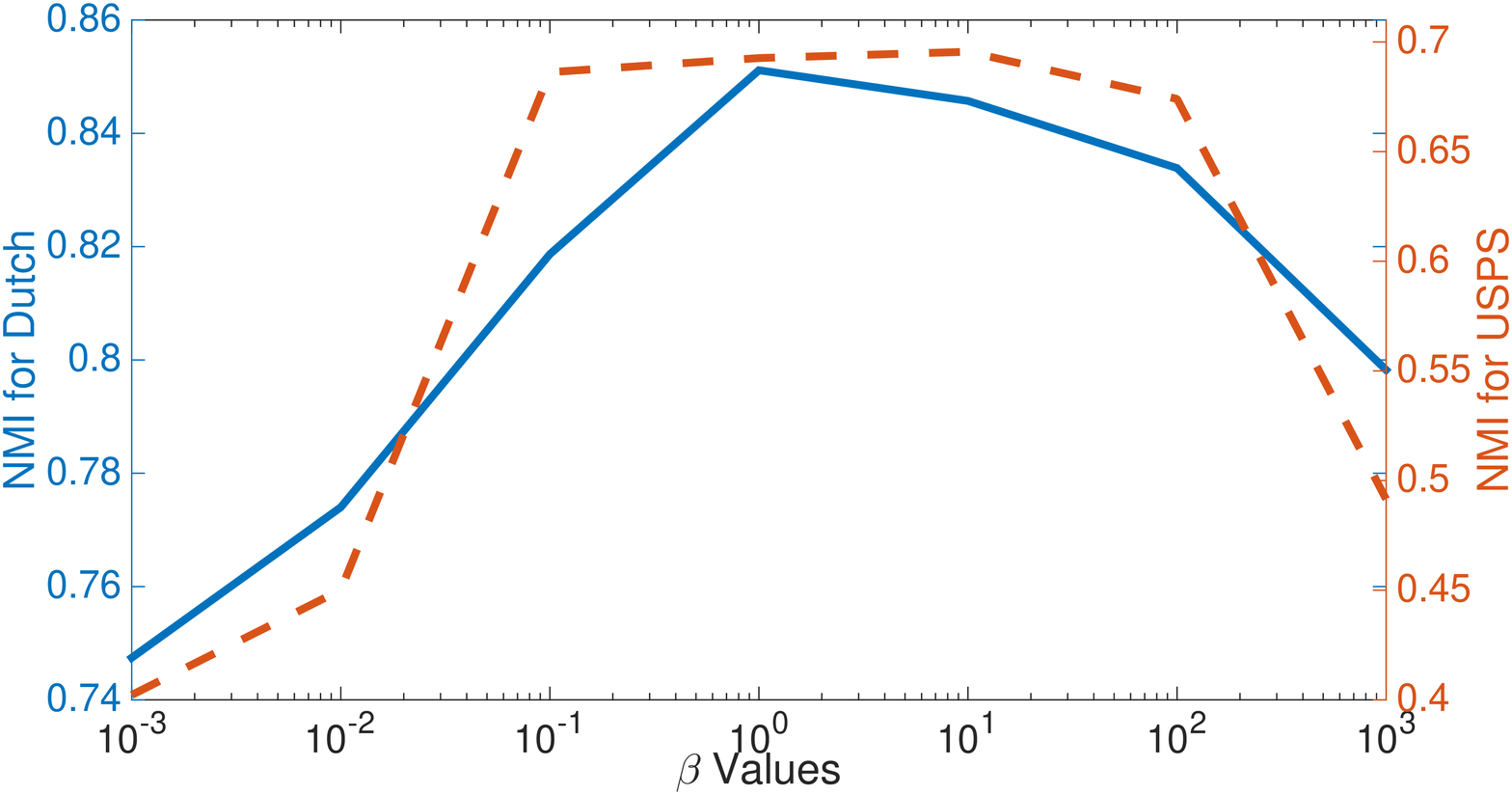}
		\caption{\small NMI v.s. $ \beta^{(i,j)} $ on Dutch-USPS.}
		\label{fig:beta}
	\end{minipage}
\end{figure}

There are two sets of parameters in the proposed MMC method:
$\{\alpha_i^k\} $, the relative importance of view $i$ in source $k$,
and $\{\beta^{(i,j)}\}$, the weight of the discrepancy penalty between source $i$ and $j$.
Here we explore the influence of the view importance weights and the discrepancy penalty weights.
We first fix $\{\beta^{(i,j)}\}$ to be 1, and run the proposed MMC method with various $\{\alpha_i^k\}$ values ($10^{-3}$ to $10^3$).
We then fix $\{\alpha_i^{k}\}$ to be 0.1, and run the proposed MMC method with various $\{\beta^{(i,j)}\}$ values ($10^{-3}$ to $10^3$).
Due to the limit of space, we only report the results on Dutch-USPS data with 60\% known mapping in Fig.~\ref{fig:alpha} and Fig.~\ref{fig:beta}.
From Fig.~\ref{fig:alpha}, we can see that the performance is stable with $ \alpha_i^k $ smaller than 100, and
the best performance is achieved when $ \alpha_i^k $ is around 0.1.
In Fig.~\ref{fig:beta}, The performance is stable with $ \beta^{(i,j)} $ between  0.1 and 100.
The best performance is achieved when $ \beta^{(i,j)} $ is near 1.
\section{Discussion}
In this section,
we aim at analyzing MMC more in detail
in order to answer the following four questions:\\
(1) How does the difficulty of the clustering problem affect the performance of these methods?\\
(2) How does percentage of known mappings between sources affect the performance of MMC?\\
(3) How good is the inferred similarity mapping?\\
(4) How fast does MMC converge?

To show the performance for clustering problem with different difficulties,
we apply MMC to Dutch-USPS data but with different number of clusters (2 to 10).
The difficulty of the clustering problem increases as the number of clusters increases.
The results are shown in Table \ref{tab:clust_1}.
The percentage of known mappings is also set to 60\%.
The NMIs for both sources are reported in separate rows (MMC(D) for Dutch and MMC(U) for USPS).

From Table \ref{tab:clust_1}, we can observe that
as the number of clusters increases (the difficulty of the problem increases),
the performance for all of the methods decrease.
The proposed MMC outperforms other comparison methods
in almost all cases with one exception.
CoReg-M outperforms MMC on USPS when the cluster number is 4.
However, the proposed MMC achieved the second best performance in that case.

To answer the second question,
we apply MMC on both Dutch-USPS data and English-Translation data
with various percentages of known mapping between 30\% to 100 \% (10\% interval).
The results are shown in Tables \ref{tab:v1} and \ref{tab:v2}.
It is worth noting that Sym-NMF, CoReg-S and MultiNMF-S
do not utilize the instance mapping across sources.
Thus, the performance of these three methods remain the same for different percentages.

In Table \ref{tab:v1} and Table \ref{tab:v2}, the proposed MMC outperforms the other comparison
methods for both sources in almost all of the different parameter settings.
It is important to notice that even with 100\% mapping available,
the proposed MMC is still better than other multi-view clustering methods.
This is because MMC will treat views within one source as a cohesive set while other multi-view clustering algorithms treat
the views from different sources equally.
\begin{table*}[t]
	\centering
	\caption{\small NMI for Dutch-USPS data with various percentages of known mapping}
	\label{tab:v1}
	\begin{tabular}{|c|c|c|c|c|c|c|c|c|}
		\hline
		Method&30\% known&40\% known&50\% known&60\% known&70\% known&80\% known&90\% known &100\% known\\
		\hline
		\hline
		Concat(D)	&	0.5515	&	0.5616	&	0.5631	&	0.5734	&	0.5770	&	0.5887	&	0.6064	&	0.6128	\\ \hline
		Sym-NMF(D)	&	0.7778	&	0.7778	&	0.7778	&	0.7778	&	0.7778	&	0.7778	&	0.7778	&	0.7778	\\ \hline
		MultiNMF-S(D)	&	0.5382	&	0.5382	&	0.5382	&	0.5382	&	0.5382	&	0.5382	&	0.5382	&	0.5382	\\ \hline
		MultiNMF-M(D)	&	0.6081	&	0.6676	&	0.7031	&	0.7585	&	0.8040	&	0.8130	&	0.7799	&	0.8356	\\ \hline
		CoReg-S(D)	&	0.7503	&	0.7503	&	0.7503	&	0.7503	&	0.7503	&	0.7503	&	0.7503	&	0.7503	\\ \hline
		CoReg-M(D)	&	0.7827	&	0.7861	&	0.7825	&	0.7886	&	0.8038	&	0.8343	&	0.8492	&	0.8596	\\ \hline
		CGC(D)	&	\textbf{0.7947}	&	0.7891	&	0.8019	&	0.7851	&	0.7929	&	0.7840	&	0.7878	&	0.8003	\\ \hline
		MMC(D)	&	0.7931	&	\textbf{0.7903}	&	\textbf{0.8177}	&	\textbf{0.8248}	&	\textbf{0.8371}	&	\textbf{0.8537}	&	\textbf{0.8611}	&	\textbf{0.8746}	\\ \hline \hline
		Concat(U)	&	0.3067	&	0.3231	&	0.3623	&	0.3916	&	0.4354	&	0.4789	&	0.5350	&	0.6128	\\ \hline
		Sym-NMF(U)	&	0.3005	&	0.3005	&	0.3005	&	0.3005	&	0.3005	&	0.3005	&	0.3005	&	0.3005	\\ \hline
		MultiNMF-S(U)	&	0.401	&	0.4010	&	0.4010	&	0.4010	&	0.4010	&	0.4010	&	0.4010	&	0.4010	\\ \hline
		MultiNMF-M(U)	&	0.3468	&	0.3611	&	0.5029	&	0.4700	&	0.6007	&	0.6083	&	0.7011	&	0.7816	\\ \hline
		CoReg-S(U)	&	0.4044	&	0.4044	&	0.4044	&	0.4044	&	0.4044	&	0.4044	&	0.4044	&	0.4044	\\ \hline
		CoReg-M(U)	&	0.3527	&	0.4094	&	0.4607	&	0.5257	&	0.5808	&	0.6642	&	0.7541	&	0.8564	\\ \hline
		CGC(U)	&	0.2968	&	0.2902	&	0.2795	&	0.2780	&	0.2958	&	0.2882	&	0.2758	&	0.2898	\\ \hline
		MMC(U)	&	\textbf{0.496}	&	\textbf{0.5463}	&	\textbf{0.6039}	&\textbf{	0.6348}	&	\textbf{0.6862}	&	\textbf{0.7587}	&	\textbf{0.8262}	&	\textbf{0.8684}	\\ \hline
	\end{tabular}
\end{table*}
\begin{table*}[t]
	\centering
	\caption{\small  NMI for English-Translation data with various percentages of known mapping}
	\label{tab:v2}
	\begin{tabular}{|c|c|c|c|c|c|c|c|c|}
		\hline
		Method&30\% known&40\% known&50\% known&60\% known&70\% known&80\% known &90\% known& 100\% known\\
		\hline
		\hline
		Concat(E)& 0.1144 &	0.1334 &	0.1617 &	0.1914 &	0.2037 &	0.2488 &	0.2495	& 0.2498\\					
		\hline
		Sym-NMF(E)& 0.2783 &	0.2783 &	0.2783 &	0.2783 &	0.2783 &	0.2783 &	0.2783 &	0.2783\\ 
		\hline
		MultiNMF-S(E)	& 0.3413 &	0.3413 &	0.3413 &	0.3413 &	0.3413 &	0.3413 &	0.3413 &	0.3413\\ 
		\hline
		MultiNMF-M(E) & 0.3253 &	0.3289 &	0.3339 &	0.3420 &	\textbf{0.3689} &	0.3531 &	0.3523 &	0.3508\\
		\hline
		CoReg-S(E)	& 0.3381 &	0.3381 &	0.3381 &	0.3381 &	0.3381 &	0.3381 &	0.3381 &	0.3381\\ 
		\hline
		CoReg-M(E) & 0.2060 &	0.2088 &	0.2098 &	0.2187 &	0.2194 &	0.2193 &	0.2184 &	0.2182\\
		\hline
		CGC(E)& 0.2388 &	0.2550 &	0.2656 &	0.2636 &	0.2707 &	0.2680 &	0.2737 &	0.2740\\
		\hline
		MMC(E)& \textbf{0.3436} &	\textbf{0.3485} &	\textbf{0.3571 }&	\textbf{0.3528} &	0.3576 &	\textbf{0.3595} &	\textbf{0.3558} &	\textbf{0.3637}\\
		\hline
		\hline
		Concat(T) & 0.1044 &	0.1124 &	0.1397 &	0.1621 &	0.1685 &	0.2098 &	0.2072	& 0.2198\\
		\hline
		Sym-NMF(T)& 0.1527 & 0.1527 & 0.1527 & 0.1527 & 0.1527 & 0.1527 & 0.1527 & 0.1527 \\
		\hline
		MultiNMF-S(T)	& 0.2708 &	0.2708 &	0.2708 &	0.2708 &	0.2708 &	0.2708 &	0.2708 &	0.2708\\ 
		\hline
		MultiNMF-M(T) & 0.1945 &	0.2063 &	0.2146 &	0.2164 &	0.2146 &	0.2223 &	0.2541 &	0.2581\\
		\hline
		CoReg-S(T)	& 0.2874 & 0.2874& 0.2874& 0.2874& 0.2874& 0.2874& 0.2874& 0.2874\\ 
		\hline
		CoReg-M(T) & 0.2160 &	0.2146 &	0.2178 &	0.2198 &	0.2171 &	0.2213 &	0.2210 &	0.2299\\
		\hline
		CGC(T)& 0.2288 &	0.2450 &	0.2556 &	0.2536 &	0.2607 &	0.2580 &	0.2637 &	0.2640\\
		\hline
		MMC(T)& \textbf{0.3028 }&	\textbf{0.3075} &	\textbf{0.3072} &	\textbf{0.3073} &	\textbf{0.3090} &	\textbf{0.3144} &	\textbf{0.3196} &	\textbf{0.3230}\\
		\hline
	\end{tabular}
\end{table*}

From the results, we can conclude that MMC works for various percentages of known mapping across sources.
The reason why MMC performs better is not only because it appreciates the cohesiveness of the views, but also for every iteration,
MMC tries to infer the instance similarity mapping between different sources.
Although the instance similarity mapping is not as the same as the instance mapping,
it provides extra information about the partially known instance mapping.
Thus the inferred instance similarity mappings will help improve clustering in the next iteration.

To show how good the inference of similarity mapping is,
we
perform another set of experiments to measure the quality of the inferred similarity mappings among those not-aligned instances.
For each not-aligned instance, we select the most similar instance mapped by the similarity mapping.
Then we check if the two instances are in the same class.
We test the accuracy for different number of clusters on Dutch-USPS data.
Here the percentage of known instance mapping is set to 60\%.
We reported the number of instances that are not aligned by the known mapping,
and the number of class matches by the inference of similarity mappings in Table \ref{tab:inference}.
\begin{table}
	\centering
	\caption{\small The inference accuarcy under differnet cluster numbers on Dutch-USPS}
	\label{tab:inference}
	\begin{adjustbox}{max width=0.8\columnwidth}
	\begin{tabular}{|c|c|c|c|c|c|}
		\hline
		&$ c = 2 $ & $ c = 4 $ & $ c = 6 $ & $ c = 8 $ & $ c = 10 $\\
		\hline
		\hline
		instances & 160 & 320 & 480 & 640 & 800\\
		\hline
		matches & 113 & 259 & 349 & 429 & 522\\
		\hline
		accuracy & 0.7063 & 0.8094 & 0.7271 & 0.6703 & 0.6525\\
		\hline
	\end{tabular}
	\end{adjustbox}
\end{table}
\begin{table}
	\centering
	\caption{\small The number of iterations until converge}
	\label{tab:converge}
	\begin{adjustbox}{max width=0.95\columnwidth}
	\begin{tabular}{|c|c|c|c|c|c|c|c|}
		\hline
		\% Known&$ 30\% $ & $ 40\% $ & $ 50\% $ & $ 60\% $ & $ 70\% $& $ 80\% $& $ 90\% $\\
		\hline
		\hline
		\# iter & 14 & 14 & 15 & 18 & 19 & 21 & 17\\
		\hline
	\end{tabular}
	\end{adjustbox}
\end{table}

\begin{figure}[t]
	\centering
	\includegraphics[width=0.85\columnwidth]{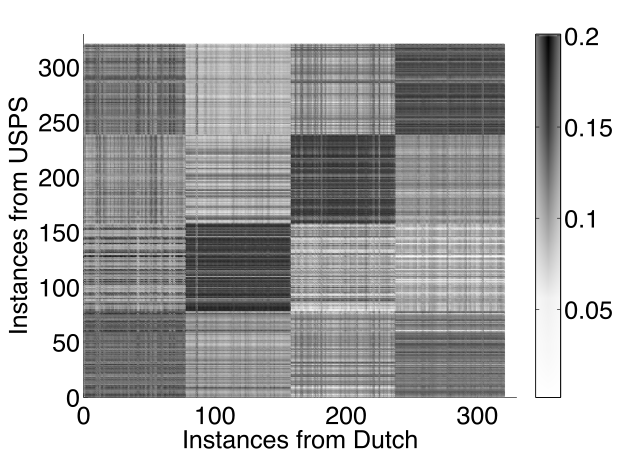}
	\caption{\small Scatter plot of the inferred similarity mapping for Dutch-USPS data with four clusters.}
	\label{fig:inference}
\end{figure}

From Table \ref{tab:inference},
we can clearly observe that when the number of clusters is 4,
the inference of similarity mapping can get an accuracy as high as 0.8094.
As the number of clusters increases, the accuracy of the inference drops.
However, even with the number of clusters being as high as 10,
we still get an accuracy of 0.6525 for the inference.
To have a better understanding of the inferred similarity mapping,
we plot the similarity mapping among the unmapped instances from the Dutch-USPS data with four clusters and 
60\% known mapping in Fig.~\ref{fig:inference}.
The instances are sorted by the class label for both sources
(the first 80 instances belongs to class 1,
the second 80 instances belong to class 2, etc,.)
The X axis indicates the instances in Dutch,
while the Y axis indicates the instances in USPS.
From the figure, we can clearly see that
there are four dark squares of width 80 on the diagonal line,
which indicates that the same class of instances are more likely to be mapped together.

To show how fast MMC converges, we report the number of outer iterations until convergence for Dutch-USPS data in Table~\ref{tab:converge}.
From the table we can see that the method converges fast (less than 20 iterations).

\section{Related Works}
Multi-view learning \cite{Blum_co_training,MVC_co_reg,Nigam_co_training,LongYZ08},
is proposed to learn from instances which have multiple
representations in different feature space.
For example, \cite{MVC} developed and studied partitioning and agglomerative,
hierarchical multi-view clustering algorithms for text data.
\cite{MVC_co_reg,MVC_co_training} are among the first works proposed to solve
clustering problem via spectral projection.
\cite{sdm2013_liu} proposed to solve multi-view clustering by joint non-negative matrix factorization.
\cite{trivedi2010multiview,ShaoMVC,PVC,DBLP:conf/pkdd/ShaoHY15} are among the first works to solve the multi-view clustering with partial/incomplete views.
However, none of the previous multi-view clustering methods can
deal with incomplete and partial known mapping between sources/views.
Further more, All the previous methods fail to treat the views within one source as a cohesive unit.

Consensus clustering \cite{consensus_clustering,li08consensus} is also related to the proposed MMC framework.
It deals with the situation in which a number of different clustering results have been obtained
for a particular dataset and it is desired to find a single consensus clustering
which is a better fit in some sense than the existing ones.
\cite{goder08consensus} gives a report about
consensus clustering algorithms comparison and refinement.
\cite{LockD13} proposes a bayesian consensus clustering method.
However, consensus clustering aims to find a single consensus clustering from fully mapped clustering solutions.
None of the previous methods works for the incomplete and partially unknown mappings between the instances.

\section{Conclusion}
This paper is the first to investigate the problem of clustering with multiple sources and multiple views.
The proposed MMC framework treats views in the same source as a cohesive group for clustering
by learning consensus latent feature matrices from the views within one source.
It also incorporates multiple sources
by using cross-source discrepancy penalty to enhance the clustering performance.
MMC also uses the learned latent features to infer the cross-source unknown similarity mapping,
which in turn will help improve the performance of clustering.
Extensive experiments conducted on three groups of real-world data sets show the effectiveness of MMC comparing with other
state-of-arts methods.
\end{document}